\documentclass[11pt,a4paper]{article}
\usepackage[hyperref]{acl2021}
\usepackage{times}
\usepackage{latexsym}

\usepackage{microtype}

\usepackage{bbding}
\usepackage{pifont}
\usepackage[nointegrals]{wasysym}
\usepackage{amssymb}
\usepackage{amsfonts} 
\usepackage{amsmath}
\usepackage{graphicx}
\usepackage{multirow}
\usepackage{amsmath}
\usepackage{hyperref}       
\usepackage{url} 
\usepackage{booktabs}       
\usepackage[ruled,vlined]{algorithm2e}
\usepackage{subfigure}

\usepackage{float}

\DeclareMathOperator*{\argmax}{argmax}
\DeclareMathOperator*{\argmin}{argmin}
\newcommand{\tabincell}[2]{\begin{tabular}{@{}#1@{}}#2\end{tabular}}

\aclfinalcopy 

\title{LV-BERT: Exploiting Layer Variety for BERT}

\author{Weihao Yu \\
  National University \\
  of Singapore \\
  \texttt{weihaoyu6@gmail.com} \\\And
  Zihang Jiang \\
  National University \\
  of Singapore \\
  \texttt{jzihang@u.nus.edu} \\ \And
  Fei Chen \\
  Huawei Noah’s \\
  Ark Lab \\
  \texttt{chen.f@huawei.com} \\ \AND
  Qibin Hou \\
  National University \\
  of Singapore \\
  \texttt{andrewhoux@gmail.com} \\ \And
  Jiashi Feng \\
  National University \\
  of Singapore \\
  \texttt{elefjia@nus.edu.sg}
  }

\date{}

\begin{document}
\maketitle
\begin{abstract}
Modern pre-trained language models are mostly built upon backbones stacking self-attention and feed-forward layers in an interleaved order.
In this paper, beyond this stereotyped layer pattern, we aim to improve pre-trained models 
by exploiting \textit{layer variety} from two aspects: the layer type set and the layer order.
Specifically, besides the original self-attention and feed-forward layers,
we introduce convolution into the layer type set, which is experimentally found beneficial to pre-trained models.
Furthermore, beyond the original interleaved order, we explore more layer orders to discover more powerful architectures.
However, the introduced layer variety leads to a large architecture space of more than billions of candidates, 
while training a single candidate model from scratch already requires huge computation cost, making it not affordable to search such a space by directly training large amounts of candidate models.
To solve this problem, we first pre-train a supernet from which the weights of all candidate models can be inherited,
and then adopt an evolutionary algorithm guided by pre-training accuracy to find the optimal architecture. 
Extensive experiments show that LV-BERT model obtained by our method outperforms BERT and its variants on various downstream tasks.
For example, LV-BERT-small achieves 79.8 on the GLUE testing set, 1.8 higher than
the strong baseline ELECTRA-small. \footnote{\url{https://github.com/yuweihao/LV-BERT}}

\end{abstract}

\section{Introduction}

\begin{figure}[tb]
    \begin{subfigure}
    \centering
    \includegraphics[width=2.3in]{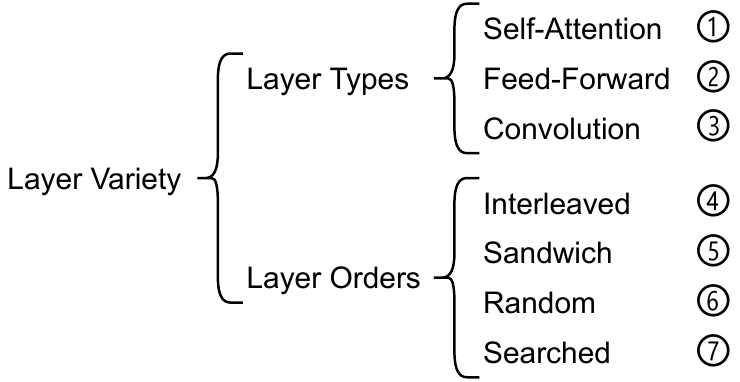}
    \vspace{-0.5cm}
    \caption*{(a)}
    \end{subfigure}
    \begin{subfigure}
    \centering
    \includegraphics[width=3in]{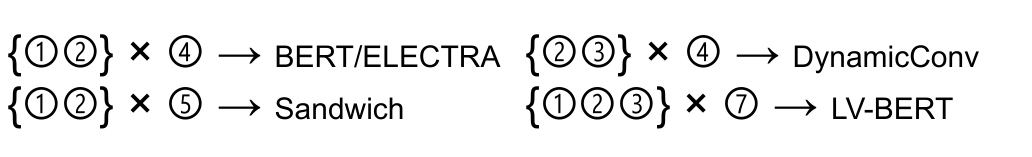}
    \vspace{-0.7cm}
    \caption*{(b)}
    \end{subfigure}
    \begin{subfigure}
    \centering
    \includegraphics[width=2.8in]{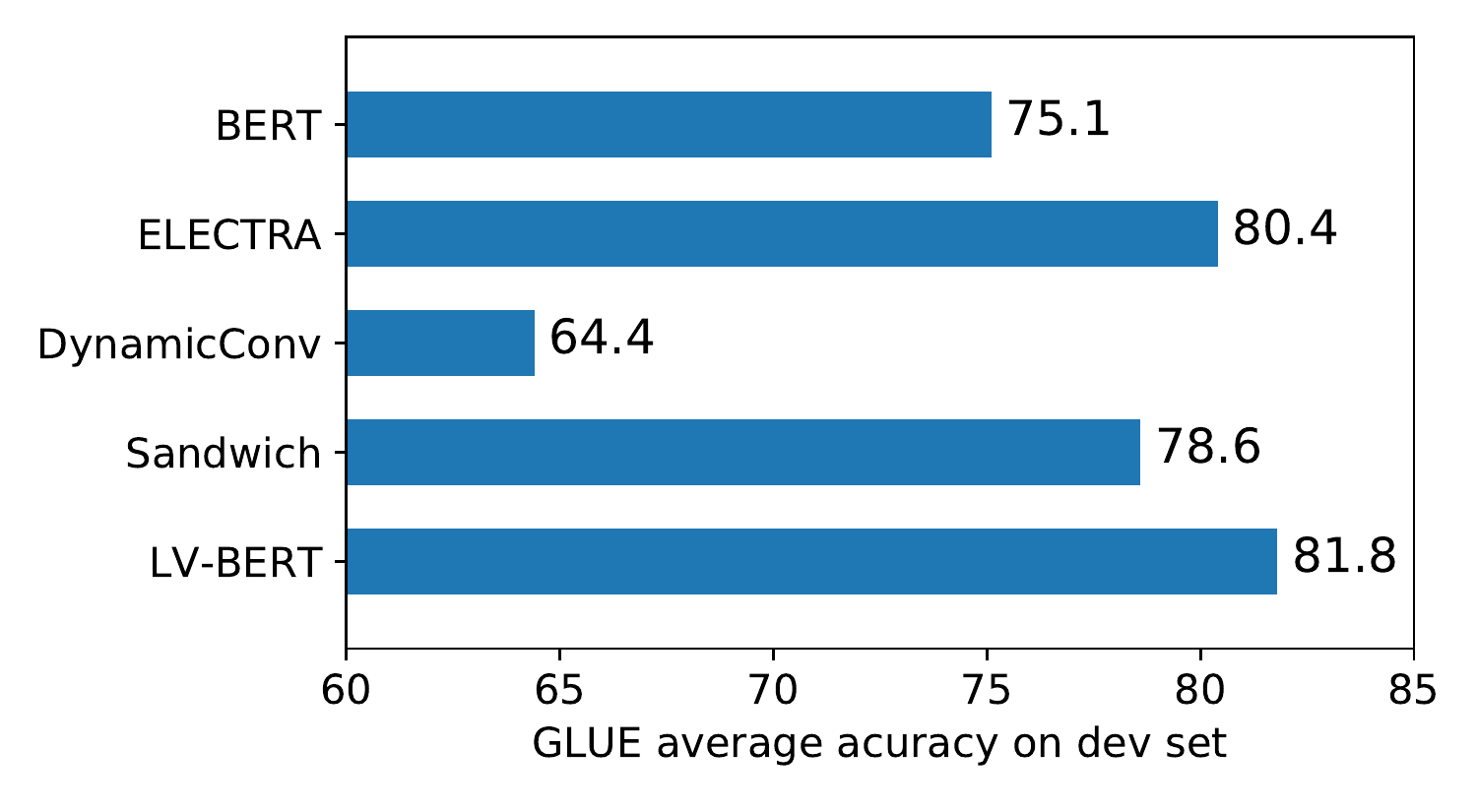}
    \vspace{-0.7cm}
    \caption*{(c)}
    \end{subfigure}
    \caption{
    (a) Illustration of layer variety. This concept consists of two aspects: layer type and layer order. 
    (b) Different models represented by layer variety.
    (c) Performance of different models with hidden size of 256 on GLUE \citep{glue} development set. 
    Except BERT pre-trained with the Masked Language Modeling objective \citep{bert}, the other models are pre-trained with Replaced Token Detection objective \citep{electra} to save computation cost.
    }
    \label{fig:first-figure}
 \end{figure}

In recent years, pre-trained language models, such as the representative BERT \citep{bert} and GPT-3 \citep{gpt3}, have gained great success in natural language processing tasks~\citep{elmo, gpt, xlnet, electra}.
The backbone architectures of these models mostly adopt a stereotyped layer pattern, in which the
\textit{self-attention} and \textit{feed-forward} layers are arrayed in an interleaved order \citep{transformer}. 
However, there is no evidence supporting that this layer pattern is optimal \citep{sandwich}. 
We then consider a straightforward and interesting question: 
Could we change the layer pattern to improve pre-trained models?
We attempt to answer this question by exploiting more \textit{layer variety} from two aspects, 
as shown in Figure \ref{fig:first-figure}(a): the layer type set and the layer order.

We first consider the layer types.
In previous pre-trained language models, the most widely-used layer set contains the self-attention layer for capturing global information and the feed-forward layer
for non-linear transformation.
However, some recent works have unveiled that some self-attention heads in pre-trained models tend to learn local dependencies due to the inherent property of natural language~\citep{kovaleva2019revealing, Brunner2020On, convbert}, incurring computation redundancy for capturing local information.
In contrast, convolution is a local operator \citep{lecun1998gradient, krizhevsky2012imagenet, simonyan2015very, he2016deep} and has shown effectiveness on extracting local information for language models \citep{zeng-etal-2014-relation, kim-2014-convolutional, kalchbrenner-etal-2014-convolutional, dynaconv, lite-transformer, convbert}. 
Thus, we propose to augment the layer set by including convolution for local information extraction.

For layer orders, most of the existing pre-trained models adopt an interleaved order to
arrange the different types of layers. 
Differently, \citet{sandwich} presented the sandwich order, \textit{i.e.}, stacking consecutive 
self-attention and feed-forward layers at the bottom and top, respectively, while keeping the interleaved order in the middle. 
It has been shown that the sandwich order can bring improvement on language modeling task, indicating the layer order contributes to model performance.
However, \citet{sandwich} did not show the generalization capability of this order to other tasks.
There is still a large room for exploring more effective orders for pre-trained models.
We show the different layer variety designs of existing models in Figure \ref{fig:first-figure}(b), including BERT \citep{bert}/ELECTRA \citep{electra}, DynamicConv \citep{dynaconv} and Sandwich \citep{sandwich}.
Their performance is summarized in Figure \ref{fig:first-figure}(c).
It can be seen that layer variety significantly influences model performance. 
We thus claim it is necessary to investigate layer variety for promoting pre-trained models. 
However, to perform such investigation for a common model backbone, \textit{e.g.}, with 24 layers, we need to evaluate performance of every candidate within an architecture space of $3^{24} \approx 2.8 \times 10^{11}$ candidates.
Pre-training a single language model already needs to consume a large amount of computation, \textit{e.g.}, 2400 P100 GPU days for pre-training BERT \citep{lin2020multi}. 
It is barely affordable to pre-train such a large amount of model candidates from scratch. 
To reduce the computation cost, inspired by recent works on Neural Architecture Search (NAS) \citep{spos, cai2019once}, 
we construct a supernet according to the layer variety discussed above and pre-train it with Masked Language Modeling (MLM) \citep{bert} objective.
After obtaining the  pre-trained supernet, we develop an evolutionary algorithm guided by MLM evaluation accuracy to search an effective architecture with specific layer variety.
We call the resulted model LV-BERT. 
Extensive experiments show that LV-BERT outperforms BERT and its variants.
The contributions of our paper are two-fold.
Firstly, to the best of our knowledge, this work is the first 
to exploit layer variety w.r.t. both layer types and orders 
for pre-trained language models. 
We found convolutions and layer orders both benefit pre-trained model performance.
We hope our observations would facilitate the development of
pre-trained lauguage models.
Secondly, our obtained LV-BERT shows superiority over BERT and its variants. For example, LV-BERT-small achieves 79.8 on GLUE testing set, 1.8 higher than the baseline ELECTRA-small \cite{electra}.

\section{Related Work}
\paragraph{Pre-trained Language Models}
Pre-trained language models have achieved great success and promoted the development of NLP techniques. 
Instead of separate word representation \citep{word2vec1, word2vec2}, \citet{mccann2017learned} and \citet{Peters:2018} propose CoVe and ELMo respectively which both utilize LSTM \citep{hochreiter1997long} to generate contextualized word representations. 
Later, \citet{gpt} introduce GPT that changes the backbone to transformers where self-attention and feed-forward layers are arrayed interleavedly.
They also propose generative pre-training objectives.
BERT \citep{bert} continues to use the same layer set and order for backbone but employs different pre-training objectives, \textit{i.e.}, Masked Language Modeling and Next Sentence Prediction.
Then more works introduce new effective pre-training objectives, like Generalized Autoregressive Pretraining \citep{xlnet}, Span Boundary Objective \cite{joshi2019spanbert} and Replaced Token Detection \citep{electra}. 
Besides designing pre-training objectives, some other works try to extend BERT by incorporating knowledge \citep{zhang2019ernie, peters2019knowledge, liu2020k, Xiong2020Pretrained} or with multiple languages \citep{huang2019unicoder, conneau2019cross, chi2019cross}.
These works utilize the stereotyped layer pattern, which is unnecessarily optimal \citep{sandwich}, inspiring us to further investigate more layer variety to improve pre-trained models. 
To the best of our knowledge, we are the first to exploit layer variety from both the layer type set and the layer order for pre-trained language models.

\paragraph{Neural Architecture Search}
Manually designing neural architecture is a time-consuming and error-prone process \citep{nas-sv}. 
To solve this, many neural architecture search algorithms are proposed.
Pioneering works utilize reinforcement learning \citep{zoph17neural, baker17designing} or evolutionary algorithm \citep{real2017large} to sample architecture candidates and train them from scratch, which demand huge computation that ordinary researchers can not afford. 
To reduce computation cost, recent methods \citep{pham2018efficient,liu2018darts,xie2018snas,brock2018smash,cai2018proxylessnas,bender2018understanding,wu2019fbnet, spos} adopt a \textit{weight sharing} strategy that a supernet subsuming all architectures is trained only once and all architecture candidates can inherit their weights from the supernet.
Despite the boom of NAS research, most works focus on computer vision tasks \citep{chen2019detnas, ghiasi2019fpn, liu2019auto}, while NAS on NLP is not fully investigated. 
Recently, \citet{so2019evolved} and \citet{wang2020hat} search architectures of transformers for translation tasks.
\citet{chen2020adabert} leverage differentiable neural architecture to automatically compress BERT with task-oriented knowledge distillation for specific tasks. 
\citet{zhu2020autorc} utilize architecture search to improve models based on pre-trained BERT for the relation classification task. 
However, these methods only focus on specific tasks or the fine-tuning phase. 
Besides, \citet{khetan2020schubert} employ pre-training loss to help prune BERT, but their method can not find new architectures. 
Different from them, our work is the first to use NAS to help explore new architectures in a pre-training scenario for general language understanding. 

\section{Method}
\label{sec:method}
\begin{figure*}[t]
     \centering
     \includegraphics[width=0.8\textwidth]{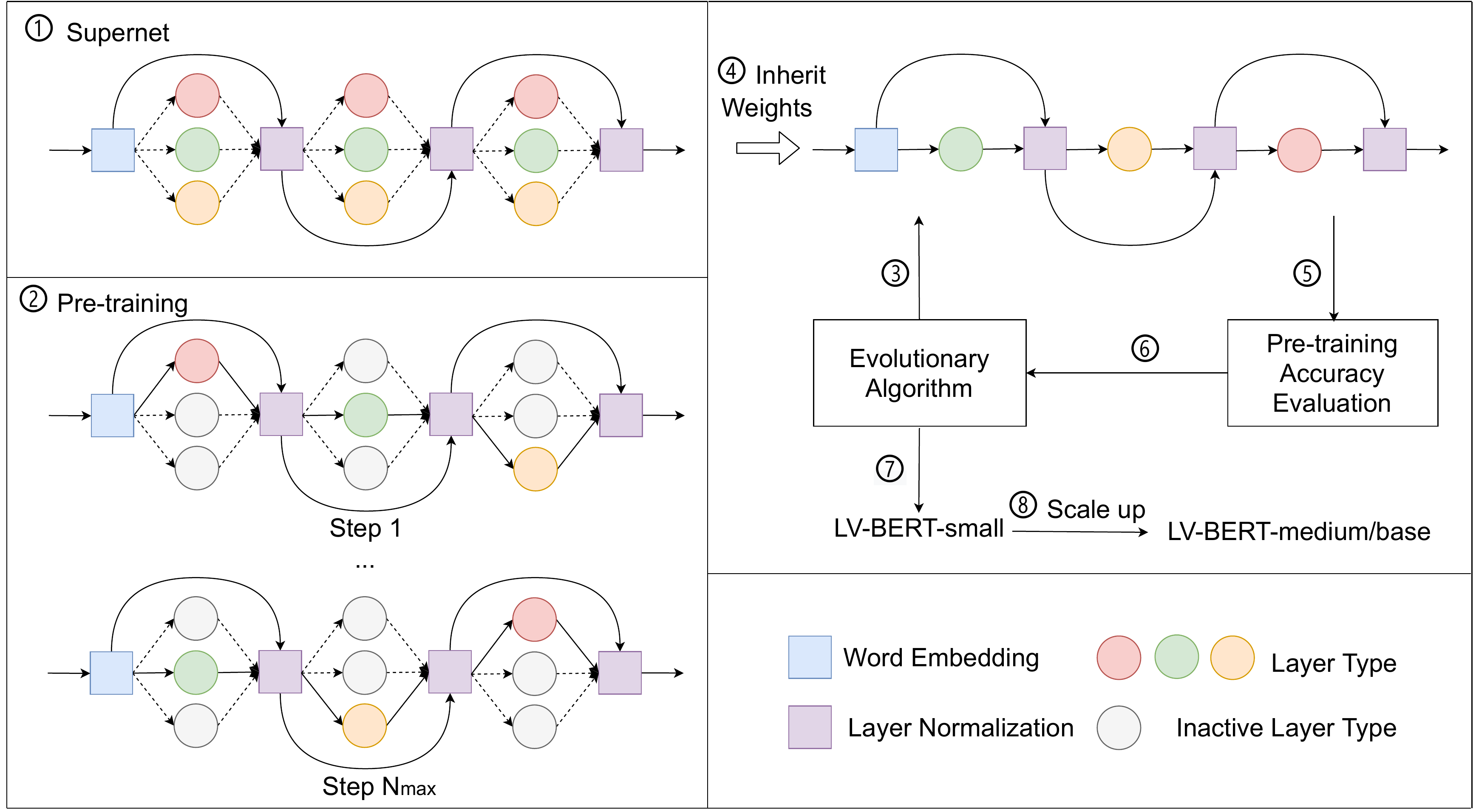}
     \caption{Overview on how to search  LV-BERT. \ding{172} Construct a supernet with small hidden size by including all types of layers at each layer. 
     \ding{173} Pre-train the supernet with Masked Language Modeling (MLM) objective \citep{bert} by only uniformly sampling one type of layer into training at each layer. \ding{174} Apply evolutionary algorithm  to produce candidate models. 
     \ding{175} The candidate models inherit their weights from the supernet. 
     \ding{176} The candidate models with inherited weights are directly evaluated with pre-training MLM accuracy on validation set. 
     \ding{177} The accuracy is used to guide the evolutionary algorithm for generating new candidate models.
     \ding{178} After $T$ iterations, the candidate with best pre-training accuracy is output as LV-BERT-small. 
     \ding{179} LV-BERT-small can be scaled up to LV-BERT-medium/base with larger hidden size. 
     } 
     \label{fig:steam}
 \end{figure*}

An overview of our approach is shown in Figure \ref{fig:steam}. We first define the layer variety to introduce a large architecture search space, and then pre-train a supernet subsuming all candidate architectures, followed by an evolutionary algorithm guided by pre-training MLM \citep{bert} accuracy to search an effective model. 
In what follows, we will give detailed descriptions.
 
\subsection{Layer Variety}

As shown in Figure \ref{fig:first-figure}(a), the proposed layer variety contains 
two aspects: layer type and layer order,  both of which are important for the performance of  pre-trained models but not exploited before. 

\paragraph{Layer Type}

The layer type set of current BERT-like models consists of 
self-attention for information communication and feed-forward for non-linear transformation.
However, as a global operator, self-attention needs to take as input all tokens to compute attention weights for each token,
which is inefficient in capturing local information \citep{lite-transformer, convbert}.
We notice that convolution \citep{lecun1998gradient, krizhevsky2012imagenet}, as a local operator, has been successfully applied in language models \citep{zeng-etal-2014-relation, kim-2014-convolutional, kalchbrenner-etal-2014-convolutional, dynaconv, lite-transformer, convbert}.
A typical example is the dynamic convolution \citep{dynaconv} for machine translation, language modeling and summarization.
Therefore, we augment the layer type set by introducing dynamic convolution as a new layer type. 
The layer set considered in this work thus contains three types of layers,
\begin{equation}
    \mathcal{L}^\mathrm{type} = \{ L^\mathrm{SA}, L^\mathrm{FF}, L^\mathrm{DC}\},
\end{equation}
where the set elements denote self-attention, feed-forward and dynamic convolution layers respectively.
See Appendix for more detailed formulation description on them. 

\paragraph{Layer Order}

The other variety aspect is layer order. 
The most widely-used order for pre-trained models is 
the interleaved order \citep{transformer, bert}.
For a model with 24 layers, the interleaved order can 
be expressed by the following list,
\begin{equation}
    [L_1^\mathrm{SA}, L_2^\mathrm{FF}, L_3^\mathrm{SA}, L_4^\mathrm{FF}, ..., L_{23}^\mathrm{SA}, L_{24}^\mathrm{FF}].
\end{equation}
Similarly, the sandwich order \citep{sandwich} can be expressed as
\begin{equation}
\begin{split}
    [& L_1^\mathrm{SA}, L_2^\mathrm{SA}, ..., L_5^\mathrm{SA}, \\
    & L_6^\mathrm{SA}, L_7^\mathrm{FF}, L_8^\mathrm{SA}, L_9^\mathrm{FF}, ..., L_{18}^\mathrm{SA}, L_{19}^\mathrm{FF},\\ 
    &L_{20}^{FF}, L_{21}^{FF}, ..., L_{24}^\mathrm{FF}].
\end{split}
\end{equation}
Beyond the above manually designed orders,
we take advantage of neural architecture search to identify more effective layer orders for pre-trained models. 
The order to be discovered can be expressed as
\begin{equation}
    [L_1, L_2, ..., L_i, ..., L_N],
\end{equation}
where $L_i \in \mathcal{L}^\mathrm{type}$ and $N$ is the number of layers. Here, $N$ is set to 24, following common practice.

\begin{algorithm}[ht]
\textbf{Input}: $W_\mathcal{A}$: supernet weights; $P$: population size; $D_{\mathrm{val}}$: pre-training validation set; $T$: \# iteration; 
$N^\mathrm{cro}$: \# crossover; $N^\mathrm{mut}$: \# mutation; $p$: mutation probability; $k$: \# top candidates   for crossover and mutation

\textbf{Output}: $a^*$: the architecture with the best pre-trianing MLM validation accuracy\\

$S_0 := \mathrm{Init}(P$); \textcolor{gray}{// Randomly generate $P$ architecture candidates}\\
$S^\mathrm{topk} := \emptyset$; \textcolor{gray}{// The set of top $k$ candidates} \\
\For{$i = 1 : T$}{
    $S_{i-1}^\mathrm{MLM} := \emptyset$; \\
    \For{$a\ \mathrm{in} \ S_{i-1}$}{
    $\mathrm{MLM}_\mathrm{val}^a := \mathrm{Inference}(\mathcal{N}(a, W_\mathcal{A}(a)), D_\mathrm{val});$ \\
    $S_{i-1}^\mathrm{MLM} := S_{i-1}^\mathrm{MLM} \cup {\mathrm{MLM}_\mathrm{val}^a};$
    }
    $S^\mathrm{topk} := \mathrm{Update}(S^\mathrm{topk}, \mathrm{S}_{i-1},S^\mathrm{MLM}_{i-1}$); 
    \\
    $S^\mathrm{cro} := \mathrm{Crossover}(S^\mathrm{topk}, N^\mathrm{cro})$; 
    \\
    $S^\mathrm{mut} := \mathrm{Mutation}(S^\mathrm{topk}, N^\mathrm{mut}, p)$; 
    \\
    $S_{i} := \mathrm{S}^\mathrm{cro} \cup \mathrm{S}^\mathrm{mut}$;\\
  }
\textbf{return} $a^*=\argmax_{a \in S^\mathrm{topk}}\mathrm{MLM}_\mathrm{val}^a$; 
\caption{\label{alg:evolution} Evolutionary Search Guided by Pre-training MLM Accuracy 
}
\end{algorithm}
\subsection{Supernet}
The layer variety introduced above leads to a huge architecture space of $3^{24} \approx 2.8 \times 10^{11}$ candidate models
to be explored.
Thus, it is not affordable to pre-train every candidate model in the space from scratch  to evaluate their performance since the pre-training procedure requires huge computations.
To reduce the search computations, recent NAS works \citep{pham2018efficient, spos, cai2019once} 
exploit a weight sharing strategy.
It first trains a supernet subsuming all candidate architectures, and then each candidate architecture can inherit its weights from the trained supernet to avoid training from scratch.
Inspired by this strategy, we construct a supernet where each layer contains all types of layers, \textit{i.e.}, self-attention, feed-forward, and dynamic convolution. The supernet architecture can be expressed as
\begin{equation}
\begin{split}
    \mathcal{A} = [&\{L_1^\mathrm{SA}, L_1^\mathrm{FF}, L_1^\mathrm{DC}\}, \{L_2^\mathrm{SA}, L_2^\mathrm{FF}, L_2^\mathrm{DC}\}, ..., \\
    &\{L_N^\mathrm{SA}, L_N^\mathrm{FF}, L_N^\mathrm{DC}\}].
\end{split}
\end{equation}
Masked Language Modeling (MLM) \citep{bert} is utilized as the pre-training objective to pre-train the supernet since MLM accuracy can reflect the model performance on downstream tasks \citep{lan2019albert}. 
Most weight sharing approaches on NAS \citep{wu2019fbnet, liu2018darts} train and optimize the full supernet: 
the output of each layer is the weighted sum of all types of candidate layers.
However, it cannot guarantee the sampled single type of layer also works \citep{spos}. 

To handle this issue, we propose to randomly sample a submodel from the supernet to participate in forward and backward propagation per training step \citep{cai2018proxylessnas, spos}.
The sampled submodel architecture can be expressed as \begin{equation}
    a = [L_1, L_2, ..., L_i, ...,  L_N], 
\end{equation}
where $L_i \in \mathcal{L}^\mathrm{type} \sim U$ with uniform probability distribution $Pr = 1/3$. 
In this pre-training method, the optimized supernet weights can be expressed as
\begin{equation}
    W_\mathcal{A} = \argmin_{W}\mathbb{E}_{a\sim U(A)}[\mathcal{L}_\mathrm{pre-train}(\mathcal{N}(a, W(a)))],
\end{equation}
where $W(a)$ denotes the submodel weights inherited from the supernet, $\mathcal{N}$ means the submodel with specific architecture and weights, $\mathcal{L}_\mathrm{pre-train}$ denotes the pre-training MLM loss and $a\sim U(\mathcal{A})$ means $a$ is uniformly sampled from $\mathcal{A}$.

\subsection{Evolutionary Search}
Inspired by the recent NAS works \citep{nas-sv, nas-sv2020, spos, wang2020hat}, we adopt an evolutionary algorithm (EA) to search the model.
Previously \citet{real2017large} utilized an evolutionary method in NAS but they trained each candidate model from scratch which is costly and inefficient. 
Instead, thanks to the supernet mentioned above, we do not need to train the candidate models from scratch since their weights can be inherited from the supernet.
Next problem is how to select indicator of the candidate models
to guide the EA. 
Note that our goal is  to search a general pre-trained model to 
benefit a variety of downstream tasks instead of a specific 
task.
Traditional NAS methods \citep{chen2020adabert, zhu2020autorc} use downstream task performance as the objective to search for task-specific models. Instead, similar to the work by \citet{khetan2020schubert} that utilize pre-training loss to prune BERT, our method uses pre-training MLM accuracy to search for a unified architecture that can generalize well to different downstream tasks. 
Besides, using this accuracy, candidate models can be directly evaluated on pre-training validation set without any fine-tuning on specific tasks, which can help save computations. 

The detailed algorithm description is shown in Algorithm~\ref{alg:evolution}.
$\mathrm{Crossover}(S^\mathrm{topk}, N^\mathrm{cro})$ means the procedure to generate $N^\mathrm{cro}$ new candidate architectures that two candidate architectures randomly selected from top $k$ candidate set $S^\mathrm{topk}$ are crossed to produce a new one. 
Similarly, $\mathrm{Mutation}(S^\mathrm{topk}, N^\mathrm{mut}, p)$ denotes the procedure to generate $N^\mathrm{mut}$ new candidates that a random candidate from $S^\mathrm{topk}$ mutates its every layer choice with probability $p$ to generate a new one. 
Finally, the candidate architecture with highest pre-training validation accuracy in $S^\mathrm{topk}$ is returned as LV-BERT.
The algorithm is set with population size $P$ of 50, search iteration number $T$ of 20, crossover number $N^\mathrm{cro}$ of 25, mutation number $M^\mathrm{mut}$ of 25, mutation probability $p$ of 0.1, top candidate number $k$ of 10 for crossover and mutation.

\section{Experiments}

\begin{table*}[!h]
\centering
\small
\begin{tabular}{lccclccc}
\hline
\textbf{Model} & \multicolumn{4}{c}{\textbf{Layer Variety}} & \multicolumn{2}{c}{\textbf{Params}} & \textbf{GLUE}\\
 & DC & SA & FF & Order & Word Emb & Backbone \\
\hline
BERT-small \citep{bert} & & \checkmark & \checkmark & Interleaved & \multirow{4}{*}{3.9M} & 9.5M & 75.1 \\
ELECTRA-small \cite{electra} & & \checkmark & \checkmark & Interleaved & & 9.5M & 80.4 \\
DynamicConv-small* \cite{dynaconv}  & \checkmark & & \checkmark & Interleaved & & 9.6M & 64.4 \\
Sandwich-small* \cite{sandwich} & \checkmark & & \checkmark & Sandwich & & 9.5M & 78.6 \\
\hline
 \multirow{11}{*}{LV-BERT-small variants}
& & \checkmark & \checkmark & Random & \multirow{11}{*}{3.9M} & 9.5M & 80.8 \\
& & \checkmark & \checkmark & Randomly searched & & 9.8M & 81.1\\
& & \checkmark & \checkmark & EA searched & & 10.3M & 81.2 \\
& \checkmark &  & \checkmark & Random & & 9.6M & 64.9\\
& \checkmark &  & \checkmark & Randomly searched & & 9.6M & 65.4\\
& \checkmark &  & \checkmark & EA searched & & 9.6M & 65.7\\
& \checkmark & \checkmark &  & Random & & 6.4M & 79.7\\
& \checkmark & \checkmark &  & Randomly searched & & 6.4M & 79.9\\
& \checkmark & \checkmark &  & EA searched & & 6.4M & 79.8\\
& \checkmark & \checkmark & \checkmark & Random & & 7.7M & 80.6\\
& \checkmark & \checkmark & \checkmark & Randomly searched & & 8.8M & 80.9\\
\hline
LV-BERT-small & \checkmark & \checkmark & \checkmark & EA searched & 3.9M & 8.5M & \textbf{81.8}\\
\hline
\end{tabular}
\caption{\label{tab:layer-variety}
Performance of the models with different layer types and orders on the GLUE development set. DC, SA and FF denote dynamic convolution, self-attention and feed-forward layers respectively. For each design of layer type set, ``Random" means the best order among five randomly generated ones that are estimated by training model from scratch.  ``Randomly searched" or ``EA searched" are both based on the supernet. ``Randomly searched" denotes the orders searched at random while ``EA searched" denotes ones searched by evolutionary algorithm. * denotes the methods implemented by us for language pre-training. All models are pre-trained on OpenWebText by 1M steps with sequence length 128 using ELECTRA \citep{electra} pre-training objective except BERT-small using MLM objective.
}
\end{table*}
\subsection{Datasets}
\paragraph{Pre-training Datasets}

\begin{table*}[!htbp]
\centering
\small
\setlength\tabcolsep{1.5pt}
\begin{tabular}{llccccccccccc}
\hline
\textbf{Model} & \textbf{Size} & \multicolumn{2}{c}{\textbf{Params}} & \textbf{CoLA} & \textbf{MPRC}  & \textbf{MNLI} & \textbf{SST}  & \textbf{RTE} & \textbf{QNLI} & \textbf{QQP} & \textbf{STS} & \textbf{Avg.}\\
 & & Word Emb & Backbone &  &  \\
\hline
ELECTRA \citep{electra} & Small & 3.9M & 9.5M & 56.8 & 87.4 & 78.9 & 88.3 & 68.5 & 87.9 & 88.3 & 86.8 & 80.4 \\
& Medium*   & 3.9M & 21.3M & 61.2 & 89.5 & 82.1 & 89.1 & 65.7 & 88.9 & 90.5 & 89.3 & 82.0\\
& Base* & 23.4M & 85.0M  & 64.8 & 88.5 & 85.7 & 92.6 & 76.5 & 91.7 & \textbf{91.1} & 89.9 & 85.1 \\
\hline
DynamicConv$^\dag$ \citep{dynaconv} & Small & 3.9M & 9.6M & 60.2 & 69.2 & 56.6 & 85.6 & 49.5 & 68.0 & 82.1 & 44.1 & 64.4 \\
& Medium  & 3.9M & 21.4M & 61.5 & 67.9 & 55.7 & 85.9 & 49.1 & 68.3 & 83.3 & 51.6 & 65.4 \\
& Base &  23.4M & 85.2M & 62.1 & 70.6 & 61.0 & 88.5 & 51.3 & 72.0 & 85.6 & 64.7 & 69.5 \\
\hline
Sandwich$^\dag$ \citep{sandwich} & Small & 3.9M & 9.5M & 53.2 & 87.1 & 77.5 & 88.1 & 63.9 & 86.4 & 88.3 & 84.6 & 78.6 \\
& Medium   & 3.9M & 21.3M & 55.6 & 86.2  &81.5 & 90.3 & 63.0 & 88.9 & 89.6 & 86.6 & 80.2  \\
& Base & 23.4M & 85.0M & 58.8 & 89.7 & 83.8 & 91.9 & 72.6 & 90.2 & 90.1 & 88.5 & 83.2 \\
\hline
LV-BERT 
& Small &  3.9M & 8.5M & 62.3 & 86.9 & 81.1 & 89.9 & 69.0 & 88.9 & 89.3 & 87.4 & 81.8 \\
& Medium & 3.9M & 19.0M & 64.4 & 88.0 & 82.4 & 90.5 & 68.6 & 89.4 & 90.1 & 89.7
 & 82.9 \\
& Base & 23.4M & 75.7M & \textbf{66.8} & \textbf{90.3} & \textbf{86.3} & \textbf{93.2} & \textbf{76.9} & \textbf{92.3} & 90.9 & \textbf{90.8} & \textbf{85.9} \\

\hline
\end{tabular}
\caption{\label{tab:generalization}
Performance of different models in different sizes on GLUE development set. * denotes results obtained by running official code. $^\dag$ denotes the methods implemented by us for language pre-training. All models are pre-trained on OpenWebText by 1M steps with sequence length 128 using ELECTRA \citep{electra} pre-training objective.
}
\end{table*}
\citet{bert} propose WikiBooks corpus for training BERT including English Wikipedia and BooksCorpus \citep{zhu2015aligning}. However, BooksCorpus is no longer publicly available. 
To ease reproduction, we train models on OpenWebText \citep{Gokaslan2019OpenWeb} that is open-sourced and of similar size with the corpus used by BERT. When pre-training the supernet, we leave 2\% data as our validation set for evolutionary search. 

\paragraph{Fine-tuning Datasets}
To compare our model with other pre-trained models, we fine-tune LV-BERT on GLUE \citep{glue}, including various tasks for general language understanding, and SQuAD 1.1/2.0 \citep{rajpurkar2016squad, rajpurkar2018know} for question answering. See Appendix for more details of all tasks.

\subsection{Implementation Details}
\paragraph{Model Size}
Similar to \citet{bert}, \citet{electra} and \citet{convbert},  we define different model sizes, \textit{i.e.}, ``small", ``medium" and ``base", with the same layer number of 24 but different hidden sizes of 256, 384, and 768, respectively. The detailed hyperparameters are shown in Appendix.

\paragraph{Pre-training Supernet}
To reduce training cost, we construct the supernet only in small size.
Since the layer number of models in medium and base sizes are the same as that of the small-sized one, the obtained architecture of LV-BERT-small can be  easily scaled up to the ones of medium and base sizes. 
We use Adam \citep{adam} to pre-train the supernet with MLM loss \citep{bert} , learning rate of 2e-4, batch size of 128, max sequence length of 128 and pre-training step number of 2 million. See Appendix for more details. 

\paragraph{Evaluation Setup} 
To compare with other pre-trained models, we pre-train the searched LV-BERT architecture for 1M steps from scratch on the OpenWebText \citep{Gokaslan2019OpenWeb} using Replaced Token Detection \citep{electra} since it can save computation cost.
We fine-tune LV-BERT on GLUE \citep{glue} and SQuAD \citep{rajpurkar2016squad, rajpurkar2018know} downstream tasks with most hyperparameters the same as those of ELECTRA \citep{electra} for fair comparison.
For GLUE tasks, the evaluation metrics are Matthews correlation for CoLA, Spearman correlation for STS, and accuracy for other tasks, which are averaged to get GLUE score. 
We utilize evaluation metrics of Exact-Match (EM) and F1 for SQuAD 1.1/2.0. 
Some of the fine-tuning datasets are small, and consequently, the results may vary substantially for different random seeds. 
Similar to ELECTRA \citep{electra}, we report the median of 10 fine-tuning runs from the same pre-trained model for each result.
See Appendix for more evaluation details.

\subsection{Ablation Study}
\paragraph{Layer Variety}
Various models are constructed with different layer variety designs, and their results on GLUE development set are shown in Table \ref{tab:layer-variety}. 
For the layer types, if only two layer types are provided, selecting self-attention and feed-forward yields the best result, which can always achieve performance higher than 80 under different search methods.
With only dynamic convolution and feed-forward, the performance drops dramatically to around 65. Surprisingly, without feed-forward, the layer set of dynamic convolution and self-attention can still achieve relatively good score, near 80. 
When using all the three layer types, we can obtain the best 81.8 score, 1.4 higher than the strong baseline ELECTRA (80.4) and 0.6 higher than the model searched with only self-attention and feed-forward (81.2). 
This indicates that it is effective to augment the layer type set by including convolution to extract local information for pre-trained models.

For layer orders, with the same layer types, the models with either EA or randomly searched orders perform better than those with randomly sampled orders, reflecting
the importance of investigating layer orders.
For example, with the same layer types of self-attention and feed-forward, the EA searched model obtains 81.2 score, improving BERT/ELECTRA by 6.1/0.8 as well as Sandwich by 2.6.

\paragraph{Search Method}
\label{para:search-method}
\begin{figure}[t]
     \centering
     \includegraphics[width=2.8in]{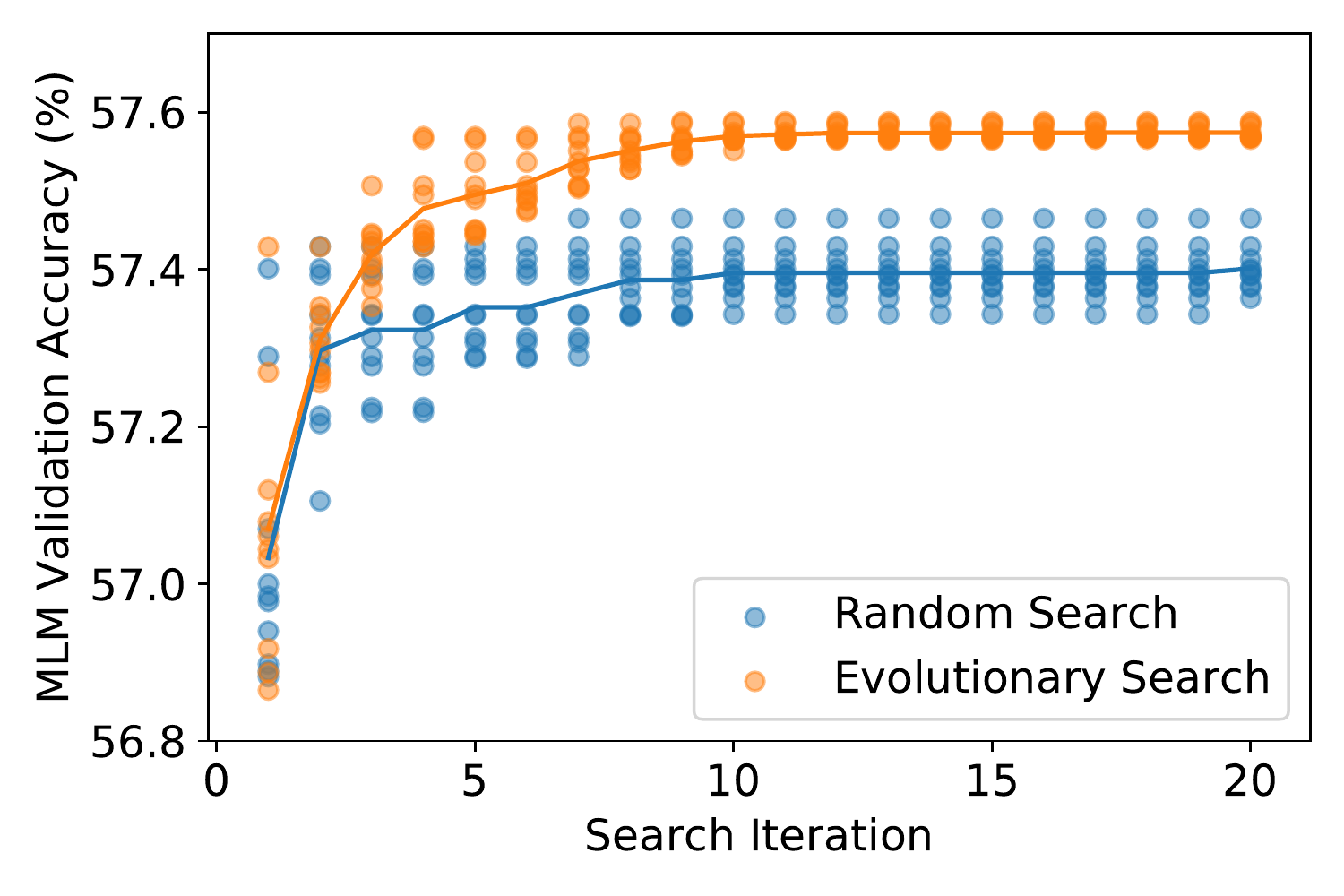}
     \caption{The pre-training MLM validation accuracy comparison between random search and evolutionary search with the layer set of all three types of layers. Blue and yellow dots denote the accuracy of top 10 candidates for each method respectively, while the plots mean their averages.
     } 
     \label{fig:scatter}
 \end{figure}

Table \ref{tab:layer-variety}  shows the results with different search methods.
``Random" means for each design of layer type set, the order is the best one among 5 randomly generated orders that are estimated by training models from scratch.
``Randomly searched" and ``EA searched" are both supernet-based methods, in which the weights of candidate models are inherited from the supernet. 
``Randomly searched" produces candidate models at random for estimation while ``EA searched" generates candidate models with evolutionary algorithm guided by the pre-training MLM accuracy.
With the same layer types, EA searched orders are generally better than randomly searched ones while the randomly searched ones are generally better than random ones. 
Figure \ref{fig:scatter} plots the pre-trianing MLM evaluation accuracy over search iterations with both random and evolutionary search methods. It shows that the accuracy of evolutionary search is obviously higher than that of random search, demonstrating the effectiveness of evolutionary search.

\begin{table*}[!h]
\centering
\small
\setlength\tabcolsep{2pt}
\begin{tabular}{l l c c c c c c c c c c c}
\hline
\textbf{Model} & \textbf{Train FLOPs} & \textbf{Params} & \textbf{CoLA} & \textbf{MPRC}  & \textbf{MNLI} & \textbf{SST}  & \textbf{RTE} & \textbf{QNLI} & \textbf{QQP} & \textbf{STS} & \textbf{Avg.}\\
\hline
TinyBERT* \citep{tinybert} & 6.4e19+ (54x+)& 15M & 51.1 & 82.6 & 84.6 & 93.1 & 70.0 & 90.4 & 89.1 & 83.7 & 80.6 \\
MobileBERT* \citep{mobilebert} & 6.4e19+ (54x+) & 25M & 51.1 & 84.5 & 84.3 & 92.6 & 70.4 & 91.6 & 88.3 & 84.8 & 81.0 \\
\hline
ELECTRA-small \citep{electra} & 1.4e18 (1.2x) & 14M & 54.6 & 83.7 & 79.7 & 89.1 & 60.8 & 87.7 & 88.0 & 80.2 & 78.0 \\
GPT \citep{gpt} & 4.0e19 (33x) & 117M & 45.4 & 75.7 & 82.1 & 91.3 & 56.0 & 88.1 & 88.5 & 80.0 & 75.9 \\
BERT-base \citep{bert} & 6.4e19 (54x) & 110M & 52.1 & 84.8 & 84.6 & 93.5 & 66.4 & 90.5 & 89.2 & 85.8 & 80.9 \\
ELECTRA-base \citep{electra} & 6.4e19 (54x) & 110M & 59.7 & 86.7 & 85.8 & 93.4 & 73.1 & \textbf{92.7} & 89.1 & 87.7 & 83.5 \\
\hline
LV-BERT-small & 1.2e18 (1x)$^\dag$ & 13M & 57.2 & 84.1 & 81.0 & 90.4 & 64.6 & 88.9 & 88.2 & 83.8 & 79.8 \\
LV-BERT-medium & 3.1e18 (2.6x)$^\dag$ & 23M & 60.1 & 85.0 & 82.0 & 91.4 & 67.6 & 89.7 & 88.9 & 85.9 & 81.3 \\
LV-BERT-base & 1.8e19 (15x)$^\dag$ & 100M & \textbf{64.0} & \textbf{87.9} & \textbf{86.4} & \textbf{94.7} & \textbf{77.0} & 92.6 & \textbf{89.5} & \textbf{88.8} & \textbf{85.1}
\\
\hline
\end{tabular}
\caption{\label{tab:glue}
Performance of models with similar size on GLUE testing set. * denotes knowledge distillation methods that rely on large pre-trained teacher models and are orthogonal to other methods. $^\dag$ We set the sequence length as 128 for pre-training to save computation although it hurts the performance.
}
\end{table*}

\begin{table*}[!h]
\centering
\small
\begin{tabular}{l l c c c c c}
\hline
\textbf{Model} & \textbf{Train FLOPs} & \textbf{Params} & \multicolumn{2}{c}{\textbf{SQuAD 1.1}} & \multicolumn{2}{c}{\textbf{SQuAD 2.0}} \\
 & & & EM & F1 & EM & F1 \\
\hline
DistillBERT* \citep{sanh2019distilbert} & 6.4e19+ (54x+) & 52M & 71.8 & 81.2 & 60.6 & 64.1 \\
TinyBERT* \citep{tinybert} & 6.4e19+ (54x+) & 15M & 72.7 & 82.1 & 65.3 & 68.8 \\
MobileBERT* \citep{mobilebert} & 6.4e19+ (54x+) & 25M & 83.4 & 90.3 & 77.6 & 80.2 \\
\hline
ELECTRA-small$^\dag$ \citep{electra} & 1.4e18 (1.2x) & 14M & 74.3 & 81.8 & 66.8 & 69.4 \\
BERT-base \citep{bert} & 6.4e19 (54x) & 110M & 80.7 & 88.4 & 74.2 & 77.1 \\
ELECTRA-base \citep{electra} & 6.4e19 (54x) & 110M & 84.5 & \textbf{90.8} & 80.5 & 83.3 \\
\hline
LV-BERT-small & 1.2e18 (1x)$^\ddag$ & 13M & 77.1 & 84.1 & 71.0 & 73.7 \\
LV-BERT-medium & 3.1e18 (2.6x)$^\ddag$ & 23M & 79.6 & 86.4 & 74.9 & 77.5\\
LV-BERT-base & 1.8e19 (15x)$^\ddag$ & 100M & \textbf{84.8} & \textbf{90.8} & \textbf{80.9} & \textbf{83.7} \\
\hline
\end{tabular}
\caption{\label{tab:squad}
Performance of models with similar model size on SQuAD 1.1/2.0 development set. * denotes knowledge distillation methods that rely on large pre-trained teacher models and are orthogonal to other methods. $^\dag$ denotes results obtained by running official code. $^\ddag$ We set the sequence length as 128 for pre-training to save computation although it hurts the performance.
}
\end{table*}

\subsection{LV-BERT Architecture}
As shown in Table \ref{tab:layer-variety}, LV-BERT achieves the best performance. Its architecture is 
\begin{equation}
\begin{split}
    [&L_1^\mathrm{DC}, L_2^\mathrm{DC}, L_3^\mathrm{SA}, L_4^\mathrm{FF}, L_5^\mathrm{FF}, L_6^\mathrm{SA},\\ &L_7^\mathrm{DC}, L_8^\mathrm{FF}, L_9^\mathrm{FF}, L_{10}^\mathrm{SA}, L_{11}^\mathrm{DC}, L_{12}^\mathrm{DC}, \\
    &L_{13}^\mathrm{SA}, L_{14}^\mathrm{FF}, L_{15}^\mathrm{DC}, L_{16}^\mathrm{FF}, 
    L_{17}^\mathrm{SA}, L_{18}^\mathrm{DC}, \\
    &L_{19}^\mathrm{FF}, L_{20}^\mathrm{DC}, L_{21}^\mathrm{SA}, L_{22}^\mathrm{SA}, L_{23}^\mathrm{FF}, L_{24}^\mathrm{SA}].
\end{split}
\end{equation}
Pre-trained with MLM from scratch by 1M steps (sequence length 128) on OpenWebText, LV-BERT-small can achieve 61.2\% MLM accuracy while BERT-small is 60.4\%.
More specific architectures of the models in Table \ref{tab:layer-variety} are listed in Appendix.

When running the evolutionary method with different seeds, we see that the resulting models prefer stacking dynamic convolutions at the bottom two layers for extracting local information and self-attention at the top layer to fuse the global information. According to these observation, for ELECTRA-small, if we replace the bottom two layers with dynamic convolutions or the top layer with self-attention, the performance can be improved by 0.3 or 0.5 respectively on GLUE development set. If we replace the bottom 8 layers with manually designed ‘ccsfccsf’ (‘c’, ‘s’ and ‘f’ denote dynamic convolution, self-attention and feed-forward layers, respectively) and replace the top 8 layers with manually designed ‘ssfsssfs’ together, we observe  0.7 performance improvement. These results show that it is helpful to stack dynamic convolution at the bottom and self-attention at the top.

\subsection{Generalization to Larger Models}
We only investigate layer variety and search models in a small-sized setting to save computation cost.
It is interesting to know whether the searched models can be generalized to larger models with large hidden size.
The results are shown in Table \ref{tab:generalization}. 
For larger model size ``medium" and ``base", LV-BERTs still outperform other baseline models, demonstrating the good generalization in terms of model size.

\subsection{Comparison with State-of-the-arts}

We compare LV-BERT with state-of-the-art pre-trained models \citep{gpt, bert, electra, sanh2019distilbert, tinybert, mobilebert} on GLUE testing set and SQuAD 1.1/2.0 to show its advantages. Although more pre-training data/steps and lager model size can significantly help improve performance \citep{xlnet, roberta, lan2019albert}, due to the computation resource limit, we only pre-train our models in small/medium/base sizes for 1M steps with OpenWebText \cite{Gokaslan2019OpenWeb}. We leave evaluating models with more pre-training data/steps and larger model size for future work. We also list some knowledge distillation methods for comparison. However,  note that these methods rely on a pre-trained large teacher network and thus are orthogonal to LV-BERT and other methods.

Table \ref{tab:glue} presents the performance of LV-BERT and other pre-trained models on GLUE testing set. It shows that LV-BERT outperforms other pre-trained models with similar model size. Remarkably, LV-BERT-small/base achieve 79.8/85.1, 1.8/1.6 higher than strong baselines ELECTRA-small/base. Even compared with knowledge distillation based model MobileBERT \citep{mobilebert}, LV-BERT-medium still outperforms it by 0.3.

Since there is nearly no single model submission on SQuAD leaderboard\footnote{\href{https://rajpurkar.github.io/SQuAD-explorer/}{rajpurkar.github.io/SQuAD-explorer/}}, we only compare LV-BERT with other pre-trained models on the development sets. The results are shown in Table \ref{tab:squad}. We find that LV-BERT-small outperforms ELECTRA-small significantly, like F1 score 73.7 versus 69.4 on SQuAD 2.0. However, when we generalize LV-BERT-small to base size, the gap between LV-BERT and ELECTRA with base size is narrower than that with small size. One reason may be LV-BERT-small is searched by our method while LV-BERT-base is only generalized from LV-BERT-small with larger hidden size.

\section{Conclusion}
We are the first to exploit layer variety for improving pre-trained language models, from two aspects, \textit{i.e.}, layer types and layer orders. For layer types, we augment the layer type set by including convolution for local information extraction. For layer orders, beyond the stereotyped interleaved one, we explore more effective orders by using an evolutionary based search algorithm.
Experiment results show our obtained model LV-BERT outperforms BERT and its variants on various downstream tasks. 
\section*{Acknowledgments}
We would like to thank the anonymous reviewers for their insightful comments and suggestions.
This research/project is supported by the National Research Foundation, Singapore under its AI Singapore Programme (AISG Award No: AISG-100E/-2019-035).
Jiashi Feng was partially supported by MOE2017-T2-2-151, NUS\_ECRA\_FY17\_P08 and CRP20-2017-0006.  
The authors also thank Quanhong Fu and Jian Liang for the help to improve the technical writing aspect of this paper. 
The computational work for this article was partially performed on resources of the National Supercomputing Centre, Singapore (https://www.nscc.sg). 
Weihao Yu would like to thank TPU Research Cloud (TRC) program and Google Cloud Research Credits Program for the support of computational resources.
\bibliographystyle{acl_natbib}
\bibliography{references}
\appendix
\section{Details about Layer Types}
For a layer,   assume its  input is $I\in \mathbb{R}^{s \times c} $ and output is $O\in \mathbb{R}^{s\times c}$, where $s$ is the sequence length and $c$ is the hidden size (channel dimension). For simplicity, $c$ takes the same value for the input and output.

\paragraph{Self-Attention} The self-Attention layer, also known as multi-head self-attention \citep{transformer}, transforms the input  by three linear transformations into the key $K$, query $Q$ and value $V$ vectors respectively,
\begin{equation}
\begin{split}
    K &= \mathrm{Reshape}(IW^{\mathrm{K}} + b^{\mathrm{K}}) \\
    Q &= \mathrm{Reshape}(IW^{\mathrm{Q}} + b^\mathrm{Q}) \\
    V &= \mathrm{Reshape}(IW^{\mathrm{V}} + b^\mathrm{V}),
\end{split}
\end{equation}
where $K, Q, V \in \mathbb{R}^{h \times s \times d}$, $W^\mathrm{K}, W^\mathrm{Q}, W^\mathrm{V} \in \mathbb{R}^{c\times c}$, and $b^\mathrm{K}, b^\mathrm{Q}, b^\mathrm{V} \in \mathbb{R}^c$. Notice that $h\times d = c$ where $h$ is the number of heads and $d$ is the head dimension.  

The above $K$ and $Q$ are used to compute their similarity matrix $M$ which is then used  to generate new value $V'$:
\begin{equation}
\begin{split}
    M &= \mathrm{Softmax}(KQ^\top/\sqrt{d}) \\
    V'&= \mathrm{Reshape}(MV),
\end{split}
\end{equation}
where $M \in \mathbb{R}^{h\times s \times s}$ and $V' \in \mathbb{R}^{s \times c}$. Finally, a linear transformation is used to exchange information between different heads, followed by shortcut connection and layer normalization,
\begin{equation}
    O = \mathrm{Norm}(V'W_O+b_O + I),
\end{equation}
where $W_O \in \mathbb{R}^{c \times c}$ and $b_O \in \mathbb{R}^c$.

\paragraph{Feed-Forward}  The feed-forward layer \citep{transformer} includes two linear transformations with a non-linear activation, followed by a shortcut connection and layer normalization,
\begin{equation}
\begin{split}
    N &= \mathrm{GELU}(IW_1 + b_1) \\
    O &= \mathrm{Norm}(NW_2 + b_2 + I),
\end{split}
\end{equation}
where $W_1 \in \mathbb{R}^{c\times rc}$ and $W_2 \in \mathbb{R}^{rc\times c}$ with a ratio $r$. $\mathrm{GELU(\cdot)}$ denotes the Gaussian Error Linear Unit \citep{gelu}.

\paragraph{Dynamic Convolution} Dynamic convolution is introduced by \citet{dynaconv} to replace self-attention, which shows strong competitiveness in the tasks of machine translation, language modeling and summarization. The dynamic convolution  first uses gated linear unit (GLU) \citep{glu} to generate new representation,
\begin{equation}
    V = \mathrm{GLU}(I).
\end{equation}
Different from the vanilla dynamic convolution that directly generates dynamic kernel from  $V \in \mathbb{R}^{s \times c}$, in this work, we supplement a separate convolution \citep{mobilenets} with depthwise weights $W^{\mathrm{Dep}} \in \mathbb{R}^{k\times c}$ ($k$ is the convolution kernel size, set as 9 in this paper) and pointwise weights $W^{\mathrm{Poi}} \in \mathbb{R}^{c \times c}$ to extract local information to help the following kernel generation. Denoting the output as $S\in \mathbb{R}^{s\times c}$, the separate convolution can be formulated as 
\begin{equation}
    S_{i, :} = \left( \sum_{j=1}^{k}W^{\mathrm{Dep}}_{j, :} \cdot V_{i+j-\frac{k+1}{2}, :} \right)W^{\mathrm{Poi}}.
\end{equation}
Then the output of separate convolution is used to generate dynamic kernels,
\begin{equation}
    D = \mathrm{Softmax}(\mathrm{Reshape}(SW^{\mathrm{Dyn}})),
\end{equation}
where $W^{Dyn} \in \mathbb{R}^{c \times hk}$ and $D \in \mathbb{R}^{h \times s \times k}$. Then  lightweight convolution is applied to the reshaped $V' = \mathrm{Reshape}(V) \in \mathbb{R}^{h \times s \times d}$. The output $C \in \mathbb{R}^{h \times s \times d}$ can be expressed as
\begin{equation}
    C_{p, i, :} = \sum_{j=1}^{k}D_{p, i, j}\cdot V'_{p, i+j-\frac{k+1}{2},:}.
\end{equation} 
Finally,   $C$ is reshaped  to   $C' = \mathrm{Reshape}(C) \in \mathbb{R}^{s \times c}$ and a linear transformer is applied  to fuse the information among multiple heads, followed by a short connection and layer normalization,
\begin{equation}
    O = \mathrm{Norm}(C'W^{\mathrm{Out}} + b^{\mathrm{Out}} + I),
\end{equation}
where $W^{\mathrm{Out}} \in \mathbb{R}^{c \times c}$ and $b^{\mathrm{Out}} \in \mathbb{R}^{c}$.

\section{Details about Datasets}
\subsection{GLUE Dataset}
Introduced by \citet{glue}, General Language Understanding Evaluation (GLUE) benchmark is a collection of nine tasks for natural language understanding, where testing set labels are hidden and predictions need to be submitted to the evaluation server\footnote{\url{https://gluebenchmark.com}}.  We provide details about the GLUE tasks below.

\paragraph{CoLA}
The Corpus of Linguistic Acceptability \citep{warstadt2019neural} is a binary single-sentence classification dataset for predicting whether an sentence is grammatical or not. The samples are from books and journal articles on linguistic theory.

\paragraph{MRPC}
The Microsoft Research Paraphrase Corpus \citep{dolan-brockett-2005-automatically} is a dataset for the task to predict whether two sentences are semantically equivalent or not. It is extracted from online news sources with human annotations.

\paragraph{MNLI}
The Multi-Genre Natural Language Inference Corpus \citep{williams-etal-2018-broad} is a dataset of sentence pairs. Each pair has a premise sentence and a hypothesis sentence, requiring models to predict its relationships containing \textit{ententailment}, \textit{contradiction} or \textit{neutral}. It is from ten distinct genres of spoken and written English.

\paragraph{SST}
The Stanford Sentiment Treebank \citep{socher2013recursive} is a dataset for the task to predict whether a sentence is positive or negative in sentiment. The dataset is from movie reviews with human annotations.

\paragraph{RTE}
The Recognizing Textual Entailment (RTE) dataset is for the task to determine whether the relationship of a pair of premise and hypothesis sentences is entailment. The dataset is from several annual textual entailment challenges including RTE1 \citep{dagan2005pascal}, RTE2 \citep{haim2006second}, RTE3 \citep{giampiccolo2007third}, and RTE5 \citep{bentivogli2009fifth}.

\paragraph{QNLI}
Question Natural Language Inference is a dataset converted from The Stanford Question Answering Dataset \cite{rajpurkar2016squad}. An example is a pair of a context sentence and a question, requiring to predict whether the context sentence contains the answer to the given question.

\begin{table}
\small  
\setlength\tabcolsep{1.6pt}
\begin{center}

\begin{tabular}{l l l l l } 
\hline
 \textbf{Hyperparameter}  & \textbf{Supernet} & \textbf{Small} & \textbf{Medium} & \textbf{Base} \\
\hline
Layer number & 24 & 24 & 24 & 24 \\
Word emb. size & 128 & 128 & 128 & 768 \\
Hidden size & 256 & 256 & 384 & 768 \\
 FF inner hidden size & 1024 &    1024 & 1536 & 3072\\
 Generator size & N/A & 1/4 & 1/3 & 1/3 \\
 Head number & 4 &   4 & 6 & 12\\
 Head size & 64 &    64 & 64 & 64\\
 Learning rate & 2e-4 & 5e-4 & 5e-4 & 2e-4 \\
 Learning rate decay& Linear &   Linear & Linear & Linear\\ 
 Warmup steps& 10000   & 10000 & 10000 & 10000\\ 
  Adam $\epsilon$&  1e-6 &  1e-6 &  1e-6 & 1e-6\\
 Adam $\beta_1$&  0.9 &   0.9& 0.9 &0.9\\
 Adam $\beta_2$&  0.999 &   0.999 & 0.999 &0.999\\
 Dropout&   0.1 & 0.1 & 0.1 & 0.1\\ 
 Batch size &  128 &  128 & 128 & 256\\
 Input sequence length & 128  & 128 & 128 & 128\\
 
\hline
\end{tabular}
\end{center}
\caption{\label{tab:pretrian_config} Pre-training hyperparameters. Generator size means the multiplier for hidden size, feed-forward inner hidden size and head number to construct generator for Replaced Token Detection pre-trianing objective \citep{electra}.}
\end{table}

\begin{table}
\centering
\small
\begin{tabular}{ll}
\hline
\textbf{Hyperparameter} & \textbf{Value} \\
\hline
Learning rate & \tabincell{l}{3e-4 for small/medium size \\ 1e-4 (except 2e-4 for SQuAD) \\ for base size} \\
Adam $\epsilon$&  1e-6 \\
Adam $\beta_1$&  0.9 \\
Adam $\beta_2$&  0.999 \\
Layerwise LR decay & 0.8 for every two layers \\
Learning rate decay & Linear \\
Warmup fraction & 0.1 \\
Attention Dropout & 0.1 \\
Dropout & 0.1 \\
Weight eecay & 0.01 \\
Batch size & 32 \\
Train epochs & \tabincell{l}{10 for RTE and STS, \\ 2 for SQuAD, \\ and 3 for other tasks} \\
\hline
\end{tabular}
\caption{\label{tab:fine-tuning-config}
Fine-tuning hyperparameters.
}
\end{table}

\begin{table*}
\centering
\small
\setlength\tabcolsep{1.5pt}
\begin{tabular}{lccclcc}
\hline
\textbf{Model} & \multicolumn{4}{c}{\textbf{Layer Variety}} & \textbf{Architecture} &\textbf{GLUE}\\
 & DC & SA & FF & \multicolumn{1}{c}{Order} & & \\
\hline
BERT-small & & \checkmark & \checkmark & Interleaved & [1, 2, 1, 2, 1, 2, 1, 2, 1, 2, 1, 2, 1, 2, 1, 2, 1, 2, 1, 2, 1, 2, 1, 2] & 75.1 \\
ELECTRA-small & & \checkmark & \checkmark & Interleaved & [1, 2, 1, 2, 1, 2, 1, 2, 1, 2, 1, 2, 1, 2, 1, 2, 1, 2, 1, 2, 1, 2, 1, 2] & 80.4 \\
DynamicConv-small*  & \checkmark & & \checkmark & Interleaved & [0, 2, 0, 2, 0, 2, 0, 2, 0, 2, 0, 2, 0, 2, 0, 2, 0, 2, 0, 2, 0, 2, 0, 2] & 64.4 \\
Sandwich-small* & \checkmark & & \checkmark & Sandwich & [1, 1, 1, 1, 1, 1, 2, 1, 2, 1, 2, 1, 2, 1, 2, 1, 2, 1, 2, 2, 2, 2, 2, 2] & 78.6 \\
\hline
\multirow{11}{*}{LV-BERT-small variants}
& & \checkmark & \checkmark & Random  & [1, 1, 1, 2, 2, 1, 2, 2, 2, 1, 1, 2, 2, 2, 1, 1, 1, 1, 2, 2, 2, 2, 1, 1] & 80.8 \\
& & \checkmark & \checkmark & Randomly searched & [1, 2, 1, 1, 2, 2, 1, 2, 1, 2, 2, 1, 2, 2, 2, 1, 1, 2, 2, 1, 1, 2, 2, 1] & 81.1\\
& & \checkmark & \checkmark & EA searched & [1, 2, 1, 2, 2, 1, 2, 2, 1, 2, 2, 1, 2, 1, 2, 2, 1, 2, 2, 1, 2, 1, 2, 2] & 81.2 \\
& \checkmark &  & \checkmark & Random & [2, 0, 0, 2, 2, 0, 0, 2, 2, 2, 0, 2, 2, 0, 0, 0, 2, 2, 0, 2, 2, 0, 0, 0] & 64.9 \\
& \checkmark &  & \checkmark & Randomly searched & [2, 2, 0, 2, 2, 0, 2, 2, 2, 0, 0, 2, 2, 0, 0, 0, 0, 2, 0, 0, 2, 0, 0, 2] & 65.4 \\
& \checkmark &  & \checkmark & EA searched & [0, 0, 2, 2, 2, 2, 2, 2, 2, 2, 2, 2, 0, 0, 0, 0, 0, 0, 0, 2, 0, 0, 0, 2] & 65.7\\
& \checkmark & \checkmark &  & Random & [0, 1, 1, 0, 1, 1, 0, 0, 1, 0, 0, 0, 0, 0, 1, 1, 1, 1, 0, 0, 0, 1, 1, 1] & 79.7 \\
& \checkmark & \checkmark &  & Randomly searched & [0, 1, 0, 1, 0, 1, 0, 1, 0, 0, 0, 1, 0, 0, 0, 1, 0, 1, 0, 1, 1, 0, 1, 0] & 79.9 \\
& \checkmark & \checkmark &  & EA searched & [0, 1, 0, 0, 0, 1, 1, 0, 1, 0, 0, 1, 0, 0, 1, 0, 0, 1, 0, 0, 1, 0, 0, 1] & 79.8 \\
& \checkmark & \checkmark & \checkmark & Random & [1, 1, 0, 0, 0, 0, 1, 1, 0, 1, 1, 0, 2, 2, 2, 1, 0, 1, 0, 1, 0, 2, 2, 1] & 80.6 \\
& \checkmark & \checkmark & \checkmark & Randomly searched & [1, 1, 0, 2, 0, 1, 2, 0, 2, 2, 1, 2, 0, 1, 2, 0, 2, 2, 0, 0, 1, 1, 2, 1] & 80.9 \\
\hline
LV-BERT-small & \checkmark & \checkmark & \checkmark & EA searched & [0, 0, 1, 2, 2, 1, 0, 2, 2, 1, 0, 0, 1, 2, 0, 2, 1, 0, 2, 0, 1, 1, 2, 1] & \textbf{81.8} \\
\hline
\end{tabular}
\caption{\label{tab:architecture}
Architectures of different models and their performance on GLUE development set. In Architecture column, 0, 1, and 2 denote dynamic convolution, self-attention, and feed-forward layers respectively * denotes methods implemented by us for language pre-training.
}
\end{table*}

\paragraph{QQP}
The Quora Question Pairs dataset \cite{chen2018quora} is the dataset from Quora, requiring to determine whether a pair of questions are semantically equivalent or not.

\paragraph{STS}
The Semantic Textual Similarity Benchmark \citep{cer-etal-2017-semeval} is a collection of sentence pairs with human-annotated similarity score on a 1-5 scare.

\paragraph{WNLI}
Winograd NLI \citep{levesque2012winograd} is a small dataset for natural language inference. However, there are issues with the construction of this dataset\footnote{\url{https://gluebenchmark.com/faq}}. Therefore, this dataset is exclude in this paper for comparison as BERT \citep{bert} \textit{etc}.

\subsubsection{SQuAD dataset}
The Stanford Question Answering Dataset (\textbf{SQuAD 1.1}) \citep{rajpurkar2016squad} is a dataset of more than 100K questions which all can be answered by locating a span of text from the corresponding context passage. Besides this data, the upgraded version \textbf{SQuAD 2.0} \citep{rajpurkar2018know} supplements it with over 50K unanswerable questions.

\section{Pre-training Details}
For supernet, We pre-train it for 2M steps with hyperparameters listed in Table \ref{tab:pretrian_config}, using Masked Language Modeling (MLM) pre-training objective \citep{bert}. This objective masks 15\% input tokens that require the model to predict. The reason to use this objective is that the MLM validation accuracy can reflect the performance of models on downstream tasks \citep{lan2019albert}.

For pre-training LV-BERTs and other compared baselines like DynamicConv \citep{dynaconv} and Sandwich \citep{sandwich} from scratch, we utilize Replaced Token Detection (RTE) pre-training objective \citep{electra}. This objective employs a small generator to predict masked tokens and utilize a larger discriminator to determine predicted tokens from the generator are the same as original ones or not. RTE can help save computation cost but achieve good performance \cite{electra}. We pre-train the models for 1M steps, mostly using the same hyperparameters as ELECTRA \citep{electra}. We set the pre-training sequence length 128 that can help us save computation cost. For downstream task SQuAD 1.1/2.0 that needs longer input sequence length, we pre-train more 10\% steps with the sequence length of 512 to learn the position embedding before fine-tuning. The hyperparameters are listed in Table \ref{tab:pretrian_config}.

\section{Fine-tuning Details}
For fine-tuning on downstream tasks, most of the hyperparameters are the same as ELECTRA \citep{electra}. See Table \ref{tab:fine-tuning-config}.

\section{Searched Architectures}
The different searched architectures are listed in Table \ref{tab:architecture}.


\end{document}